\documentclass[11pt,a4paper]{article}
\usepackage[hyperref]{acl2018}
\usepackage{url}
\aclfinalcopy % Uncomment this line for the final submission
\title{The Current State of Finnish NLP}

\author{Mika Hämäläinen \\
  Faculty of Arts \\
  University of Helsinki \\
  and Rootroo Ltd \\
  {\tt mika.hamalainen@helsinki.fi} \\\And
  Khalid Alnajjar \\
  Faculty of Arts \\
  University of Helsinki \\
  and Rootroo Ltd \\
  {\tt khalid.alnajjar@helsinki.fi} \\}

\date{}

\begin{document}
\maketitle
\begin{abstract}
  There are a lot of tools and resources available for processing Finnish. In this paper, we survey recent papers focusing on Finnish NLP related to many different subcategories of NLP such as parsing, generation, semantics and speech. NLP research is conducted in many different research groups in Finland, and it is frequently the case that NLP tools and models resulting from academic research are made available for others to use on platforms such as Github.
  \begin{center} \textbf{Tiivistelmä} \end{center}
  Suomen kielen koneelliseen käsittelyyn on tarjolla paljon valmiita työkaluja ja resursseja. Tässä artikkelissa tarkastelemme viimeaikoina julkaistuja tieteellisiä artikkeleita, joissa keskitytään suomen kielen kieliteknologiaan. Tarkastelemme kieliteknologian eri alaluokkia, kuten jäsentämistä, tuottamista, semantiikkaa ja puheetta. kieliteknologista tutkimusta tehdään Suomessa monissa eri tutkimusryhmissä, ja usein akateemisen tutkimuksen tuloksena tuotetut kieliteknologian työkalut ja mallit julkaistaan muiden käytettäväksi esimerkiksi Githubissa.
\end{abstract}

\section{Introduction}

There is no doubt that, within the Uralic language family, Finnish is one of the most well-resourced languages in terms of natural language processing (NLP). This has, however, not always been the case. Currently, NLP research conducted for Finnish has started to fragment into research outputs of several different research groups, and there is no survey paper out there that would describe the current state of Finnish NLP.

We hope that this survey paper clarifies the current situation and makes it clearer for people working in the academia outside of Finnish universities or in the industry and also for students. As it has been discussed before \cite{mika-endangered}, Finnish is certainly not a low-resourced language, and our current survey further proves this point.

It is also important for researchers working on other smaller Uralic languages to see what has been done for Finnish in terms of NLP to see what the possible and meaningful directions are for further developing the resources needed. Especially since Uralic language share the same feature of rich morphology, which is something that commonly causes problems for computers.

\section{Finnish NLP}

In this section, we present a survey on the current state of Finnish NLP. We have tried to gather most of the current research on the topic, but we are certain that there are some research out there we have not been able to find. We have categorized the surveyed research outputs into parsing, generation, semantics and speech.

\subsection{Parsing}

Starting from morphology, stemming and spell checking Finnish is well supported in multiple commercial applications such as Microsoft and Google products. In the open-source world, low-level tasks such as stemming and spell checking can be conducted with Voikko\footnote{https://voikko.puimula.org/}.

Omorfi \cite{pirinen2015development}\footnote{https://github.com/flammie/omorfi} is currently the most well supported tool for morphological analysis (including lemmatization) and generation. It is an FST (finite-state transducer) based tool developed on HFST (Helsinki finite-state technology) \cite{linden2013hfst} and it works together with constraint grammar (CG) based disambiguators and syntactic parsers available in the Giellatekno \cite{Moshagen2014} repositories\footnote{https://github.com/giellalt/lang-fin/tree/main/src/cg3}.

FinnPos\footnote{https://github.com/mpsilfve/FinnPos} \cite{silfverberg2016finnpos} is another morphological tagger and lemmatizer tool based on CRF (conditional random field). There have been recently more data driven approaches focusing on Finnish \cite{silfverberg-hulden-2018-initial}.

While rule-based tradition has been strong in the past\footnote{See \citealp{pirinen2019neural} for some comparison between rules and neural networks}, there are several machine learning driven dependency parsers for Finnish, such as the statistical one\footnote{https://turkunlp.org/Finnish-dep-parser/}  \cite{haverinen2013tdt} and neural one\footnote{http://turkunlp.org/Turku-neural-parser-pipeline/} \cite{udst:turkunlp} by TurkuNLP.

Out of the aforementioned tools Omorfi (and the CG disambigator) and the machine learning based parsers are available to use through a Python package named UralicNLP\footnote{https://github.com/mikahama/uralicNLP} \footnote{https://github.com/mikahama/uralicNLP/wiki/Dependency-parsing} \cite{uralicnlp_2019}.

As Finnish data is available in several multilingual datasets, there are many multilingual approaches for parsing \cite{qi-etal-2020-stanza}\footnote{https://stanfordnlp.github.io/stanza/} \cite{spacy}\footnote{https://spacy.io/} and morphology \cite{aharoni-goldberg-2017-morphological,nicolai-yarowsky-2019-learning,silfverberg-tyers-2019-data,gronroos2020morfessor}.

The fact that spoken Finnish is very different to standard Finnish has drawn some attention in the past \cite{jauhiainen-2001-using} and recently \cite{partanen-etal-2019-dialect}. The latter leading to a Python library called Murre\footnote{https://github.com/mikahama/murre} for automatic normalization of dialectal Finnish.

Non-standard data has been an issue in digital humanities (DH) projects \cite{makela2020wrangling}, and lately there have been efforts in automatically correcting OCR errors in existing historical datasets \cite{kettunen2015keep,drobac2020optical,1647b325b0f14e76b14c259d9a24c108,duong2020unsupervised}.

Named entity recognition has also been under study with FiNER\footnote{https://github.com/Traubert/FiNer-rules/blob/master/finer-readme.md} and its recently released data \cite{ruokolainen2019finnish}. There is also another recent BERT \cite{devlin-etal-2019-bert} based approach\footnote{https://turkunlp.org/fin-ner.html} to the topic \cite{luoma2020broad}.

There have been several approaches to language detection including detection of Finnish from web corpora (see  \citealp{jauhiainenjauhiainenjauhiainen}). Similarly, native Finnish has been automatically identified from learner's Finnish \cite{malmasi2014finnish}.

In summary, parsing has been researched on different levels of language such as syntax, morphology, POS and NER tagging, and lemmatization. It has been mainly focusing on standard well-formed Finnish, although there are methods for coping with dialectal Finnish and OCR errors as well. 

\subsection{Generation}

The lowest level of natural language generation is surface realization (see \citealp{reiter1994has}), and for that there are tools such as Omorfi and Syntax Maker\footnote{https://github.com/mikahama/syntaxmaker} \cite{hamalainen-rueter-2018-development}. The latter uses Omorfi for morphological inflection while it takes care of higher level morphosyntax such as case government and agreement.

There is a strong computational creativity focus in Helsinki and it also shows in Finnish NLG, as there are several poem generators such as Keinoleino\footnote{https://github.com/mikahama/keinoleino} \cite{hamalainen2018harnessing}, Poeticus \cite{toivanen2012corpus} and others \cite{hamalainen2019generating,hamalainen2019let}. There is also an interactive poem generator tool called \textit{Runokone} (Poem Machine)\footnote{http://runokone.cs.helsinki.fi/} \cite{hamalainen2018poem}.

Recently there have been several approaches to enhancing existing news headlines \cite{alnajjar2019no,ramo2021using}. And some approaches to generating entire news articles automatically \cite{kanerva2019template,haapanen2020recycling}.

Paraphrase generation \cite{sjoblom2020paraphrase} has also become a researched topic with the availability of monolingually aligned parallel corpora \cite{creutz2018open}. There is also an approach to converting standard Finnish text into different dialects \cite{hamalainen2020automatic}.

Finnish is a typical language for machine translation tasks and it is not uncommon to see it featured in several papers that deal with multiple languages. However, there are several papers that focus on Finnish in particular \cite{hurskainen2017rule,hamalainen2019template,pirinen-2019-apertium,tiedemann-etal-2020-fiskmo}.

There is also a recent approach to dialog generation in Finnish \cite{leino2020finchat}. Also non-native language learner's errors have been corrected successfully automatically \cite{creutz2019toward}.

To summarize the approaches, there are several generators for poetry and news that benefit from the available surface realizers. Paraphrasing, dialect adaptation, dialog generation and learners' error correction are domains with some research with potential for new discoveries in the future. Machine translation gets frequently attention from different researchers. There are several more NLG tasks (see \citealt{gatt2018survey}) that have not been researched at all in Finnish, which means that there is a lot of room for more research on this topic.

\subsection{Semantics}

Vector representations of meaning have become common place in NLP and Finnish is no exception with the availability of pretrained word2vec\footnote{http://vectors.nlpl.eu/repository/} \footnote{https://bionlp.utu.fi/finnish-internet-parsebank.html} \cite{laippala2014syntactic,kutuzov2017word} and fastText\footnote{https://fasttext.cc/docs/en/pretrained-vectors.html} \cite{bojanowski2017enriching} models.

BERT models have also become available as part of the multilingual BERT model\footnote{https://github.com/google-research/bert} \cite{devlin-etal-2019-bert} or trained separately for Finnish\footnote{http://vectors.nlpl.eu/repository/} \footnote{https://github.com/TurkuNLP/FinBERT} \cite{kutuzov2017word,virtanen2019multilingual}. Even Elmo models have been made available for Finnish\footnote{https://www.clarin.si/repository/xmlui/handle/11356/1277} \cite{ulvcar2020high}.

In addition to the standard vector-based representations of meaning, there is another statistical model called SemFi\footnote{https://github.com/mikahama/uralicNLP/wiki/Semantics-(SemFi,-SemUr)} \cite{hamalainen2018extracting}. The model is a relational database that captures semantic relations of words based on their syntactic co-occurencies.

Before the era of machine learning, there were two prominent projects for modeling meaning computationally which have been translated into Finnish WordNet \cite{linden2010finnwordnet} and FrameNet \cite{linden2019finntransframe}.

With the similar ideology to the hand crafted resources, there have been several different linked data projects in Finland representing semantics in structured ontologies \cite{hyvonen2006culturesampo,nyrkko2018building,thomas2018co,koho2019warsampo}. Many of the linked data projects are available on the Linked Data Finland website\footnote{https://www.ldf.fi/}.

There is a Python library called FinMeter\footnote{https://github.com/mikahama/finmeter} \cite{hamalainen2019let} that has some higher level semantic tools for Finnish such as metaphor interpretation, word concreteness analysis and sentiment analysis. Sentiment analysis for Finnish has also been studied later on\footnote{a dataset https://github.com/Helsinki-NLP/XED} \cite{ohman2020xed,8931724,linden2020finnsentiment}. There is also research on topic modeling methods \cite{ginter2009combining, hengchen2018comparing,loukasmaki2019eduskunnan}.

Finnish is well supported by traditional representations of semantics and latest vector based models. There is a vast amount of linked data resources in a variety of domains. Higher-level semantics such as metaphor interpretation and sentiment analysis also have received their share of research interest, although there are many more questions related to pragmatics and figurative language that have not been researched, such as sarcasm detection, multi-hop reasoning and fake news detection to name a few.

\subsection{Speech}

Apart from Finnish speech being supported by companies, there are some open-source tools that can synthesize Finnish. Festival\footnote{https://www.cstr.ed.ac.uk/projects/festival/} has a Finnish voice named Suopuhe\footnote{http://urn.fi/urn:nbn:fi:lb-20140730144}, and eSpeak-ng\footnote{https://github.com/espeak-ng/espeak-ng} can even generate IPA characters for Finnish.

There are several more modern approaches to speech recognition \cite{enarvi2017automatic,varjokallio2021morphologically} and speech synthesis \cite{2be1dd96f88b49c5be7f1a06a9f2d99a,raitio2014voice}. Although, speech synthesis has not gained much interest in the recent years.

There are several approaches to analyzing speech prosody \cite{virkkunen2018prosodic,simko2020}. There is also some work on detecting different accents in spoken Finnish \cite{behravan2013foreign,behravan2015factors} and named entity recognition \cite{porjazovski2020named}.

In summary, several approaches exist for speech processing in Finnish relating to recognition, accents and prosody. However, speech synthesis has received a surprisingly small amount of attention in the recent past. With the emergence of neural models, new research on synthesis could reach to potentially interesting new contributions.

\section{Discussion and Conclusions}

In this survey, we have gathered research conducted on different aspects of NLP. We have included links to models and code implementations for most of the research papers. It has been a pleasant thing to notice that not only Finnish NLP research exists but also it is often not conducted in a closed fashion, but the actual research outputs have been made openly available for a wider community of people even outside of academia. This is crucial for any language that is relatively small, like Finnish. If Finnish academics did not release their research, there would not be many other people in the world that would produce high-quality tools for Finnish.

Digital extinction is something that many endangered languages are facing right now (see \citealt{kornai2013digital}). Therefore, it is important to ensure that NLP resources become openly available for endangered Uralic languages as well. Availability itself is not enough, however, as the resources need to be easy to find and use. Despite the fact that we have open NLP tools for Finnish, we are still far a way from a world where machines use our language fluently. Finnair's in-flight entertainment system still announces happliy: \textit{*saavumme kohteeseen Helsinki} (*we arrive in destination Helsinki) instead of expressing it correctly, \textit{saavumme Helsinkiin} (we arrive in Helsinki), Google Doc's spell checker does not recognize mostly any inflectional form with a possessive suffix and predictive text in mobile keyboards suggest overly formal normative Finnish only.

While Finnish NLP has come far in terms of academic research and tools built as a result, we as a nation are still far away from having Finnish language technology fully integrated into the systems we use every day. Many of the problems have been solved already, it is just the matter of the industry finding out about the NLP tools that are out there.

We have limited our survey to NLP tools and methods only. We know that there are a plethora of language resources available for Finnish as well. Based on our experiences, many corpora are well hidden and digging them up is a time consuming effort worthy of a separate survey paper. Unfortunately the Finnish practice of describing data on Metashare\footnote{https://metashare.csc.fi/} is very unhelpful in this respect because the metadata descriptions in the service hardly ever contain information about where to access the data, how to cite it and who the real authors are.

%\bibliography{acl2018}
\bibliography{acl2018}
\bibliographystyle{acl_natbib}

\end{document}